\documentclass{article}
\usepackage{ijcai15}
\usepackage{times}
\usepackage{subfigure}
\usepackage{graphicx}
\usepackage{amsmath}
\usepackage{amssymb}
\usepackage{multirow}

\title{Fast Constraint Propagation for Image Segmentation}
\author{Peng Han\\
School of Information, Renmin University of China, Beijing 100872, China\\
hanpeng\_ruc@ruc.edu.cn}

\begin{document}

\maketitle

\begin{abstract}
This paper presents a novel selective constraint propagation method for constrained image segmentation. In the literature, many pairwise constraint propagation methods have been developed to exploit pairwise constraints for cluster analysis. However, since most of these methods have a polynomial time complexity, they are not much suitable for segmentation of images even with a moderate size, which is actually equivalent to cluster analysis with a large data size. Considering the local homogeneousness of a natural image, we choose to perform pairwise constraint propagation only over a selected subset of pixels, but not over the whole image. Such a selective constraint propagation problem is then solved by an efficient graph-based learning algorithm. To further speed up our selective constraint propagation, we also discard those less important propagated constraints during graph-based learning. Finally, the selectively propagated constraints are exploited based on $L_1$-minimization for normalized cuts over the whole image. The experimental results demonstrate the promising performance of the proposed method for segmentation with selectively propagated constraints.
\end{abstract}

\section{Introduction}

Image segmentation is a fundamental problem in computer vision.
Despite many years of research \cite{SM00,MFTM01,CBGM02,UPH07,AMF11},
general-purpose image segmentation is still a very challenging task
because segmentation is inherently ill-posed. To improve the results
of image segmentation, much attention has been paid to constrained
image segmentation \cite{YS04,EOK07,XLS09,GA10}, where certain
constraints are initially provided for image segmentation. In this
paper, we focus on constrained image segmentation using pairwise
constraints, which can be derived from the initial labels of selected
pixels. In general, there exist two types of pairwise constraints:
must-link constraint denotes a pair of pixels belonging to the same
image region, while cannot-link constraint denotes a pair of pixels
belonging to two different image regions. In previous work
\cite{WCR01,KKM02,XNJ03,HLL06}, such weak supervisory information has
been widely used to improve the performance of machine learning and
pattern recognition \cite{lu2006iterative,lu2008comparing,lu2009generalized,lu2011latent} in many challenging tasks.

The main challenge in constrained image segmentation is how to
effectively exploit a limited number of pairwise constraints for image
segmentation. A sound solution is to perform pairwise constraint
propagation to generate more pairwise constraints. Although many
pairwise constraint propagation methods \cite{LC08,LLT08,LI10,lu2013exhaustive} have
been developed for constrained clustering \cite{KKM03,KBD05}, they
mostly have a polynomial time complexity and thus are not much
suitable for segmentation of images even with a moderate size (e.g.
$200\times 200$ pixels), which is actually equivalent to clustering
with a large data size (i.e. $N=40,000$). For constrained image
segmentation, we need to develop more efficient pairwise constraint
propagation methods, instead of directly utilizing the existing
methods like \cite{LC08,LLT08,LI10,lu2013exhaustive}. Here, it is worth noting that
even the simple assignment operation incurs a large time cost of
$O(N^2)$ if we perform pairwise constraint propagation over all the
pixels, since the number of all possible pairwise constraints is
$N(N-1)/2$. The unique choice is to propagate the initial pairwise
constraints \emph{only to a selected subset of pixels}, but not across
the whole image.

\begin{figure*}[t]
\vspace{0.02in}
\begin{center}
\includegraphics[width=0.95\textwidth]{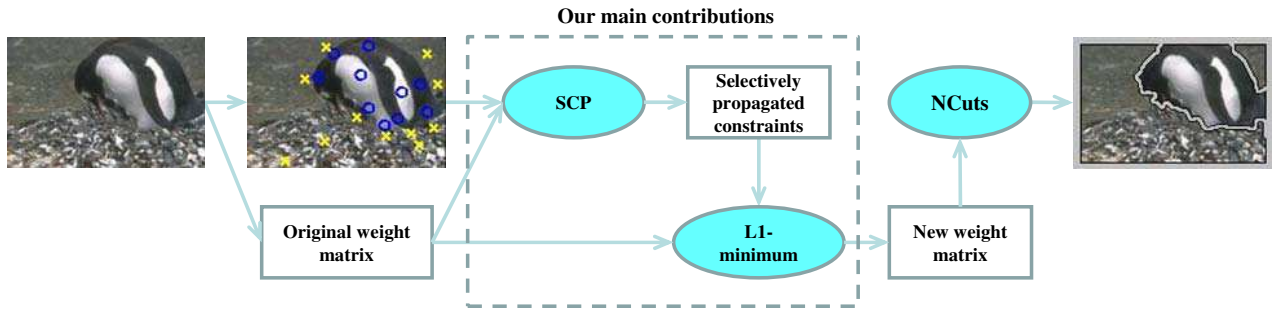}
\end{center}
\vspace{-0.12in} \caption{The flowchart of our selective constraint
propagation (SCP) for constrained image segmentation, where the
selectively propagated constraints are exploited based on
$L_1$-minimization for normalized cuts (NCuts).} \label{Fig.1}
\vspace{-0.05in}
\end{figure*}

Fortunately, the local homogeneousness of a natural image provides kind of supports for this choice, i.e., a selected subset of pixels may approximate the whole image (such downsampling is widely used in
image processing). More importantly, the selectively propagated
constraints over a selected subset of pixels are enough for achieving
a good quality of image segmentation, as verified by our later
experimental results. Hence, in this paper, we develop a selective
constraint propagation (SCP) method for constrained image
segmentation, whichpropagates the initial pairwise constraints only to a selected subset of pixels (\emph{the first meaning} of our selective constraint propagation). Although there exist different sampling
methods in statistics, we only consider random sampling for efficiency
purposes, i.e., the subset of pixels used for selective constraint
propagation are selected randomly from the whole image. In this paper,
we formulate our selective constraint propagation as a graph-based
learning problem which can be efficiently solved based on the label
propagation technique \cite{ZBLW04}. To further speed up our
algorithm, we also discard those less important propagated constraints
during iteration of graph-based learning, which is \emph{the second meaning} of our selective constraint propagation. To the best of our knowledge, we have made the first attempt to develop a selective constraint propagation method for constrained image segmentation.

Finally, the selectively propagated constraints obtained by our
selective constraint propagation are exploited to adjust the original
weight matrix based on optimization techniques, in order to ensure
that the new weight matrix is as consistent as possible with the
selectively propagated constraints. In this paper, we formulate such
weight adjustment as an \emph{$L_1$-minimization} problem
\cite{EA06,MES08,WYG09,XZY11}, which can be solved efficiently due to
the limited number of selectively propagated constraints. The obtained
new weight matrix is then applied to normalized cuts \cite{SM00} for
image segmentation. The flowchart of our selective constraint
propagation for constrained image segmentation is illustrated in
Figure~\ref{Fig.1}. Although our selective constraint propagation
method is originally designed for constrained image segmentation, it
can be readily extended to other challenging tasks (e.g., semantic
image segmentation and multi-face tracking) where only a limited
number of pairwise constraints are provided initially.

It should be noted that the present work is distinctly different from
previous work on constrained image segmentation
\cite{YS04,EOK07,XLS09,GA10}. In \cite{YS04}, only linear equality
constraints (analogous to must-link constraints) are exploited for
image segmentation based on normalized cuts. In
\cite{EOK07,XLS09,GA10}, although more types of constraints are
exploited for image segmentation, the linear inequality constraints
analogous to cannot-link constraints are completely ignored just as
\cite{YS04}. In contrast, our selective constraint propagation method
exploits both must-link and cannot-link constraints for normalized
cuts. More notably, as shown in our later experiments, our method
\emph{obviously outperforms} \cite{YS04} due to extra consideration of
cannot-link constraints for image segmentation.

The remainder of this paper is organized as follows. In Section 2, we
develop a selective constraint propagation method which propagates the
initial pairwise constraints only to a selected subset of pixels. In
Section 3, the selectively propagated constraints are further
exploited to adjust the original weight matrix based on
$L_1$-minimization for normalized cuts. Finally, Sections 4 and 5 give
the experimental results and conclusions, respectively.

\section{Selective Constraint Propagation}
\label{sect:refine}

This section presents our selective constraint propagation (SCP) in
detail. We first give our problem formulation for propagating the
initial pairwise constraints only to a selected subset of pixels from
a graph-based learning viewpoint, and then develop an efficient SCP
algorithm based on the label propagation technique \cite{ZBLW04}.

\subsection{Problem Formulation}

In this paper, our goal is to propagate the initial pairwise
constraints to a selected subset of pixels for constrained image
segmentation. To this end, we need to first select a subset of pixels
for our selective constraint propagation. Although there exist
different sampling methods in statistics, we only consider random
sampling for efficiency purposes, i.e., the subset of pixels are
selected randomly from the whole image. In the following, the problem
formulation for selective constraint propagation over the selected
subset of pixels is elaborated from a graph-based learning viewpoint.

Let $\mathcal{M}=\{(i,j): l_i=l_j,1\leq i,j\leq N\}$ denote the set of
initial must-link constraints and $\mathcal{C}=\{(i,j): l_i\neq
l_j,1\leq i,j\leq N\}$ denote the set of initial cannot-link
constraints, where $l_i$ is the region label assigned to pixel $i$ and
$N$ is the total number of pixels within an image. The set of
constrained pixels is thus denoted as $P_c=\{i:(i,j)\in
\mathcal{M}\cup\mathcal{C}, 1\leq j\leq N \}\cup \{i:(j,i)\in
\mathcal{M}\cup\mathcal{C}, 1\leq j\leq N \}$ with $N_c=|P_c|$.
Moreover, we randomly select a subset of pixels $P_s \subset
\{1,2,...,N\}$ with $N_s=|P_s|$, and then form the final selected
subset of pixels used for our selective constraint propagation as
$P_u=P_s \cup P_c$ with $N_u=|P_u|$.

In this paper, we construct a $k$-nearest neighbors ($k$-NN) graph
over all the pixels so that the normalized cuts for image segmentation
can be performed efficiently over this $k$-NN graph. Let its weight
matrix be $W=\{w(i,j)\}_{N\times N}$. We define the weight matrix
$W_u=\{w_u(i,j)\}_{N_u\times N_u}$ over the selected subset of pixels
$P_u$ as:
\begin{eqnarray}
w_u(i,j)=w(P_u(i), P_u(j)), \label{eq:knnwt}
\end{eqnarray}
where $P_u(i)$ denotes the $i$-th member of $P_u$. The normalized
Laplacian matrix is then given by
\begin{eqnarray}
L_u = I-D^{-1/2}W_uD^{-1/2}, \label{eq:lap}
\end{eqnarray}
where $I$ is an identity matrix and $D$ is a diagonal matrix with its
$i$-th diagonal entry being the sum of the $i$-th row of $W_u$.
Moreover, for convenience, we represent the two sets of initial
pairwise constraints $\mathcal{M}$ and $\mathcal{C}$ using a single
matrix $Z_u=\{z_u(i,j)\}_{N_u\times N_u}$ as follows:
\begin{eqnarray}
z_u(i,j) =\begin{cases} +1, & (P_u(i), P_u(j))\in \mathcal{M}; \\
-1, & (P_u(i), P_u(j)) \in \mathcal{C};\\
0, & \text{otherwise}.\\
\end{cases} \label{eq:initpc}
\end{eqnarray}

Based on the above notations, the problem of selective constraint
propagation over the selected subset of pixels $P_u$ can be formulated
from a graph-based learning viewpoint:
\begin{eqnarray}
&\hspace{-0.3in}&\min_{F_v,F_h}\|F_v-Z_u\|_F^2 +
\mu \mathrm{tr}(F_v^TL_uF_v) + \|F_h-Z_u\|_F^2 \nonumber \\
&\hspace{-0.3in}& \hspace{0.5in} +\mu \mathrm{tr}(F_hL_uF_h^T) + \gamma \|F_v-F_h\|_F^2,
\label{eq:intracp}
\end{eqnarray}
where $\mu$ and $\gamma$ denote the positive regularization
parameters, $||\cdot||_F$ denotes the Frobenius norm of a matrix, and
$\mathrm{tr}(\cdot)$ denotes the trace of a matrix. Here, it is worth
noting that the above problem formulation actually imposes both
\emph{vertical and horizontal} constraint propagation upon the initial
matrix $Z_u$. That is, each column (or row) of $Z_u$ can be regraded
as the initial configuration of a \emph{two-class label propagation} problem, which is formulated just the same as \cite{ZBLW04}. Moreover, in this paper, we assume that the vertical and horizontal constraint
propagation have the same importance for constrained image
segmentation.

The objective function given by Eq. (\ref{eq:intracp}) is further
discussed as follows. The first and second terms are related to the
\emph{vertical} constraint propagation, while the third and fourth
terms are related to the \emph{horizontal} constraint propagation. The
fifth term then ensures that the solutions of these types of
constraint propagation are as approximate as possible. Let $F_v^*$ and
$F_h^*$ be the best solutions of vertical and horizontal constraint
propagation, respectively. The best solution of our selective
constraint propagation is defined as:
\begin{eqnarray}
F_u^*=(F_v^*+F_h^*)/2.
\end{eqnarray}
As for the second and fourth terms, they are known as Laplacian
regularization \cite{ZGL03,ZBLW04,BNS06}, which means that $F_v$
and $F_h$  should not change too much between similar pixels. Such
Laplacian regularization has been widely used for different
graph-based learning problems in the literature.

To apply our selective constraint propagation (SCP) to constrained
image segmentation, we have to concern the following key problem:
\emph{how to solve Eq. (\ref{eq:intracp}) efficiently}. Fortunately,
due to the problem formulation from a graph-based learning viewpoint,
we can develop an efficient SCP algorithm using the label propagation
technique \cite{ZBLW04} based on $k$-NN graph over $P_u$. The proposed
SCP algorithm will be elaborated in the next subsection.

\subsection{Efficient SCP Algorithm}
\label{sect:refine:alg}

Let $\mathcal{Q}(F_v,F_h)$ denote the objective function in Eq.
(\ref{eq:intracp}). The alternate optimization technique can be
adopted to solve $\min_{F_v,F_h}\mathcal{Q}(F_v,F_h)$ as follows: 1)
Fix $F_h=F_h^*$, and perform the vertical propagation by $F_v^*=
\arg\min_{F_v} \mathcal{Q}(F_v,F_h^*)$; 2) Fix $F_v=F_v^*$, and
perform the horizontal propagation by $F_h^*=\arg\min_{F_h}
\mathcal{Q}(F_v^*,F_h)$.

\noindent \textbf{Vertical Propagation:} When $F_h$ is fixed at
$F_h^*$, the solution of $\min_{F_v}\mathcal{Q}(F_v,F_h^*)$ can be
found by solving
\begin{eqnarray}
\hspace{-0.05in}\frac{\partial \mathcal{Q}(F_v,F_h^*)}{\partial F_v} =
2(F_v-Z_u) + 2\mu L_u F_v + 2\gamma(F_v-F_h^*)= 0,\nonumber
\end{eqnarray}
which is actually equivalent to
\begin{eqnarray}
(I+\hat{\mu}L_u)F_v = (1-\beta)Z_u+\beta F_h^*, \label{eq:lev}
\end{eqnarray}
where $\hat{\mu}=\mu/(1+\gamma)$ and $\beta=\gamma/(1+\gamma)$. Since
$I+\hat{\mu}L_u$ is positive definite, the above linear equation has a
solution:
\begin{eqnarray}
F_v^*= (I+\hat{\mu}L_u)^{-1}((1-\beta)Z_u+\beta F_h^*). \label{eq:lp}
\end{eqnarray}
However, this analytical solution is not efficient at all for
constrained image segmentation, since the matrix inverse incurs a
large time cost of $O(N_u^3)$. In fact, this solution can also be
\emph{efficiently found using the label propagation technique}
\cite{ZBLW04} based on $k$-NN graph over $P_u$ (see the SCP algorithm
outlined below).

\noindent \textbf{Horizontal Propagation:} When $F_v$ is fixed at
$F_v^*$, the solution of $\min_{F_h}\mathcal{Q}(F_v^*,F_h)$ can be
found by solving
\begin{eqnarray}
\hspace{-0.05in} \frac{\partial \mathcal{Q}(F_v^*,F_h)}{\partial F_h}
= 2(F_h-Z_u) + 2\mu F_h L_u + 2\gamma(F_h-F_v^*)= 0,\nonumber
\end{eqnarray}
which is actually equivalent to
\begin{eqnarray}
F_h(I+\hat{\mu}L_u) = (1-\beta)Z_u+\beta F_v^*. \label{eq:leh}
\end{eqnarray}
This linear equation can also be efficiently solved using the label
propagation technique \cite{ZBLW04} based on $k$-NN graph, similar to
what we do for Eq.~(\ref{eq:lev}).

Since the wight matrix $W_u$ over $P_u$ is derived from the original
weight matrix $W$ of the $k$-NN graph constructed over all the pixels
according to Eq.~(\ref{eq:knnwt}), $W_u$ can be regarded as the weight
matrix of a $k$-NN graph constructed over  $P_u$. Hence, we can adopt
the label propagation technique \cite{ZBLW04} to efficiently solve
both Eq.~(\ref{eq:lev}) and Eq.~(\ref{eq:leh}). Moreover, to speed up
our selective constraint propagation, we also discard those less
important propagated constraints during both vertical and horizontal
propagation. That is, the two matrices $F_v$ and $F_h$ are forced to
become sparser and thus less computation load is needed during
iteration.

The complete SCP algorithm is outlined as follows:
\begin{description}
\item[(1)]
Compute $S_u=I-L_u$, where $L_u$ is given by Eq. (\ref{eq:lap});
\item[(2)]
Initialize $F_v(0)=0$, $F_h^*=0$, and $F_h(0)=0$;
\item[(3)]
Discard those less important propagated constraints with
$F_v(t)<\epsilon$ during the vertical propagation, where we set
$\epsilon=10^{-7}$ in this paper;
\item[(4)]
$F_v(t+1) = \alpha S_u F_v(t) +(1-\alpha) ((1-\beta)Z_u +\beta
F_h^*)$, where $\alpha=\hat{\mu}/(1+\hat{\mu}) \in (0,1)$ and
$\beta=\gamma/(1+\gamma) \in (0,1)$;
\item[(5)]
Iterate Steps (3)--(4) for the vertical propagation until convergence
at $F_v^*$;
\item[(6)]
Discard those less important propagated constraints with
$F_h(t)<\epsilon$ during the horizontal propagation;
\item[(7)]
$F_h(t+1) = \alpha F_h(t) S_u  +(1-\alpha) ((1-\beta)Z_u +\beta
F_v^*)$;
\item[(8)]
Iterate Steps (6)--(7) for the horizontal propagation until until
convergence at $F_h^*$;
\item[(9)]
Iterate Steps (3)--(8) until the stopping condition is     satisfied,
and output the solution $F_u^*=(F_v^*+F_h^*)/2$.
\end{description}

Similar to \cite{ZBLW04}, the iteration in Step (4) converges to $
F_v^*=(1-\alpha)(I-\alpha S_u)^{-1}((1-\beta)Z_u+\beta F_h^*)$, which
is equal to the solution (\ref{eq:lp}) given that $\alpha=\hat{\mu}/
(1+\hat{\mu})$ and $S_u=I-L_u$. Moreover, in our later experiments, we
find that the iterations in Steps (5), (8) and (9) generally converge
in very limited steps ($<$10). Finally, based on $k$-NN graph, our SCP
algorithm has a time cost of $O(kN_u^2)$ ($N_u\ll N$). Hence, it can
be considered to provide an efficient solution (note that even a
simple assignment operation on $F_u^*$ incurs a time cost of
$O(N_u^2)$).

\section{Constrained Image Segmentation}

In this section, we discuss how to exploit the selectively propagated
constraints stored in the output $F_u^*$ of our SCP algorithm for
image segmentation based on normalized cuts. The basic idea is to
adjust the original weight matrix $W_u$ over the selected subset of
pixels $P_u$ using these selectively propagated constraints. The
problem of such weight adjustment for normalized cuts can be
formulated as:
\begin{eqnarray}
\min_{\tilde{W}_u\geq 0}\frac{1}{2}||\tilde{W}_u-F_u^*||_F^2 +
\lambda||\tilde{W}_u-W_u||_1, \label{eq:wtadjust}
\end{eqnarray}
where $\tilde{W}_u \in R^{N_u\times N_u}$ is the new weight matrix
over $P_u$, and $\lambda$ is the regularization parameter. It is worth
noting that the new weight matrix $\tilde{W}_u$ is actually derived as
a \emph{nonnegative fusion} of $F_u^*$ and $W_u$ by solving the above
$L_1$-minimization problem. More notably, the \emph{$L_1$-norm
regularization} term $||\tilde{W}_u-W_u||_1$ can force the new weight
matrix $\tilde{W}_u$ not only to approach $W_u$ but also to become as
sparse as $W_u$, given that $W_u$ can be regarded as the weight matrix
(thus sparse) of a $k$-NN graph constructed over $P_u$.

The problem given by Eq.~(\ref{eq:wtadjust}) can be solved based on
some basic $L_1$-minimization techniques
\cite{EA06,MES08,WYG09,XZY11}. In fact, it has an explicit solution:
\begin{eqnarray}
\tilde{W}_u^*=\mathrm{soft\_thr}(F_u^*,W_u,\lambda),\label{eq:wtbest}
\end{eqnarray}
where $\mathrm{soft\_thr}(\cdot,\cdot,\lambda)$ is a soft-thresholding
function. Here, we directly define $z=\mathrm{soft\_thr}(x, y,
\lambda)$ as:
\begin{eqnarray}
z =\begin{cases} z_1=\max(x-\lambda,y), & f_1\leq f_2 \\
z_2=\max(0,\min(x+\lambda,y)), & f_1> f_2\\
\end{cases},
\end{eqnarray}
where $f_1=(z_1-x)^2 + 2\lambda|z_1-y|$ and $f_2=(z_2-x)^2 +
2\lambda|z_2-y|$. Since the $L_1$-minimization problem given by
Eq.~(\ref{eq:wtadjust}) is limited to $P_u$, finding the best new
weight matrix $\tilde{W}_u^*$ incurs a time cost of $O(N_u^2)$
($N_u\ll N$).

Once we have found the best new weight matrix $\tilde{W}_u^* =
\{\tilde{w}_u^*(i',j')\}_{N_u\times N_u}$ over the selected subset of
pixels $P_u$, we can derive the new weight matrix $\tilde{W} =
\{\tilde{w}(i,j)\}_{N\times N}$ over all the pixels from the original
weight matrix $W=\{w(i,j)\}_{N\times N}$ of the $k$-NN graph as:
\begin{eqnarray}
\tilde{w}(i,j) =\begin{cases} \tilde{w}_u^*(i',j'), & i,j\in P_u, P_u(i')=i,P_u(j')=j \\
w(i,j), & \mathrm{otherwise}\\
\end{cases}, \nonumber
\end{eqnarray}
where $P_u(i')$ denotes the $i'$-th member of $P_u$. This new weight
matrix $\tilde{W}$ over all the pixels is then applied to normalized
cuts for image segmentation.

The full algorithm for constrained image segmentation can be
summarized as follows:
\begin{description}
\item[(1)]
Generate the selectively propagated constraints using our SCP
algorithm proposed in Section~2;
\item[(2)]
Adjust the original weight matrix by exploiting the selectively
propagated constraints according to Eq.~(\ref{eq:wtbest});
\item[(3)]
Perform normalized cuts with the adjusted new weight matrix for image
segmentation.
\end{description}
As we have mentioned, Steps (1) and (2) incur a time cost of
$O(kN_u^2)$ and $O(N_u^2)$ ($N_u\ll N$), respectively. Moreover, since
the adjusted new weight matrix $\tilde{W}$ is very sparse, Step (3)
can be performed efficiently. Here, it is worth noting that the most
time-consuming component of normalized cuts (i.e., eigenvalue
decomposition) has a linear time cost when the weight matrix is very
sparse. In summary, our algorithm runs very efficiently for
constrained image segmentation.

\begin{figure*}[t]
\vspace{0.06in}
\begin{center}
\includegraphics[width=0.98\textwidth]{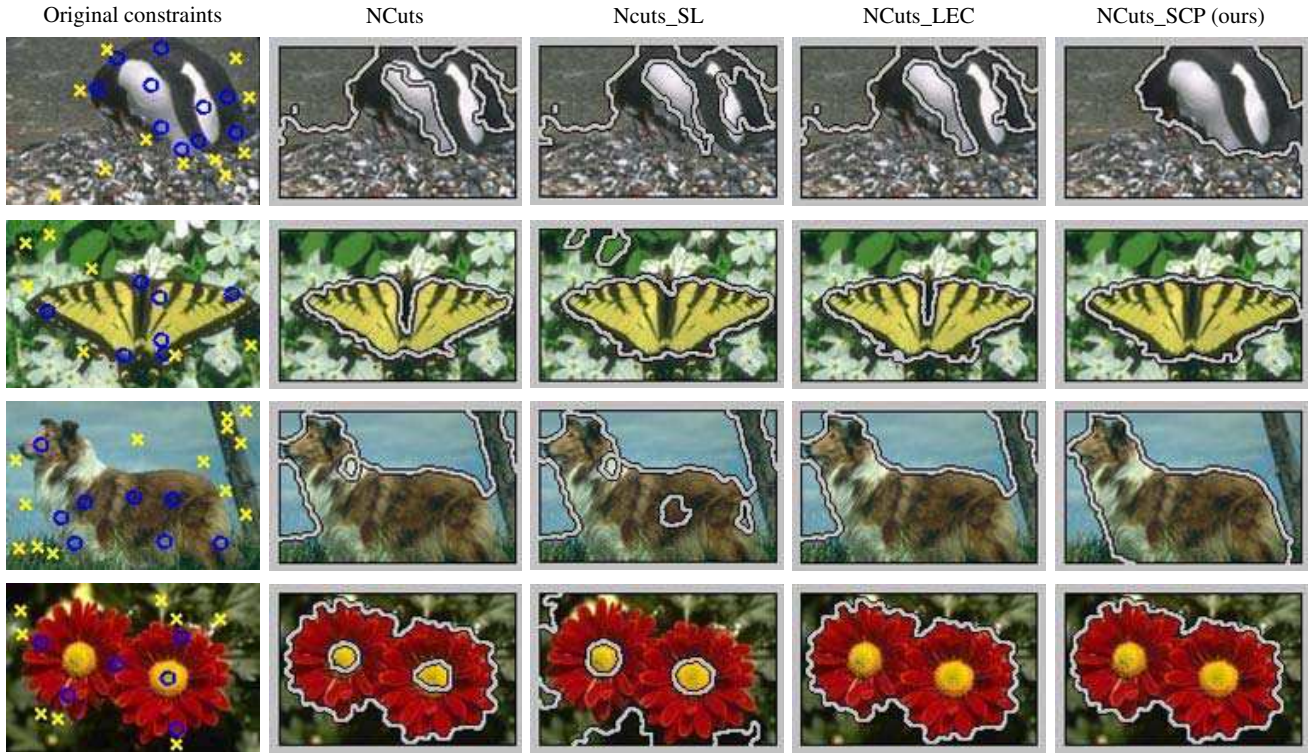}
\end{center}
\vspace{-0.1in} \caption{The results of constrained image
segmentation. A small set of labeled pixels (object and background
denoted by markers `o' and `x', respectively) are provided to infer
the initial set of linear equality constraints and pairwise
constraints.} \label{Fig.2} \vspace{-0.05in}
\end{figure*}

\section{Experimental Results}
\label{sect:exp}

In this section, our algorithm is evaluated in the task of constrained
image segmentation. We first describe the experimental setup,
including information of the feature extraction and the implementation
details. Moreover, we compare our algorithm with other closely related
methods.

\subsection{Experimental Setup}

For segmentation evaluation, we select 50 images from the Berkeley
segmentation database \cite{MFTM01} (along with ground truth
segmentations), and some sample images are shown in Figures \ref{Fig.2} and \ref{Fig.4}. It can be observed that these selected images generally have \emph{confusing backgrounds}, such as the penguin and kangaroo images. Furthermore, we consider a 6-dimensional vector of color and texture features for each pixel of an image just as \cite{CBGM02}. The three color features are the coordinates in the L*a*b* color space, which are smoothed to avoid over-segmentation arising from local color variations due to texture. The three texture features are contrast, anisotropy, and polarity, which are extracted at an automatically selected scale.

The segmentation results are measured by the adjusted Rand (AR) index
\cite{HA85} which takes values between -1 and 1, and a higher AR score
indicates that a higher percentage of pixel pairs in a test
segmentation have the same relationship (joined or separated) as in
each ground truth segmentation. We do not consider the original Rand
index \cite{Rand71,UPH07} for segmentation evaluation, since there
exists a problem with this measure \cite{HA85}. In the following, our
normalized cuts with selective constraint propagation (NCuts\_SCP) is
compared with three closely related methods: normalized cuts with
linear equality constraints (NCuts\_LEC) \cite{YS04}, normalized cuts
with spectral learning (NCuts\_SL) \cite{KKM03}, and standard
normalized cuts (NCuts) \cite{SM00}. Here, we do not make comparison
to other constraint propagation methods \cite{LC08,LLT08,LI10,lu2013exhaustive}, since
they have a polynomial time complexity and are not suitable for image segmentation.

\begin{table}[t]
\vspace{-0.05in} \caption{The average segmentation results achieved by
our NCuts\_SCP algorithm with a varied number (i.e. $N_s$) of pixels
being selected randomly.} \label{Table.1} \vspace{-0.02in}
\begin{center}
\tabcolsep0.14cm
\begin{tabular}{|c|cccccc|}
\hline
$N_s$      &  1,000 & 1,500 &  2,000  & 2,500 & 3,000  & 3,500 \\
\hline
AR index    & 0.47  &  0.48  &  0.48 &  0.50 &  0.50 & 0.50 \\
Time (sec.) &  31   &  32    &  33   &  35   &  36   & 37  \\
\hline
\end{tabular}
\end{center}
\vspace{-0.1in}
\end{table}

We randomly select a small set of labeled pixels to infer the initial
set of linear equality constraints (for NCuts\_LEC) and pairwise
constraints (for NCuts\_SCP and NCuts\_SL). Moreover, we set the
parameters of our NCuts\_SCP algorithm as:  $k=60$, $\alpha=0.9$,
$\beta=0.1$, $\epsilon=10^{-7}$, and $\lambda=0.001$. The parameters
of other closely related methods are also set their respective optimal
values.

\begin{figure}[t]
\vspace{0.06in}
\begin{center}
\includegraphics[width=0.47\textwidth]{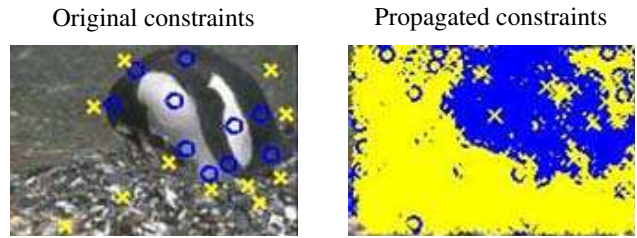}
\end{center}
\vspace{-0.05in} \caption{Illustration of the selectively propagated
constraints obtained by our NCuts\_SCP algorithm with $N_s=2,500$. The
pixels within object and background are marked by blue `o' and yellow
`x', respectively.} \label{Fig.3} \vspace{-0.00in}
\end{figure}

\begin{figure*}[t]
\vspace{0.06in}
\begin{center}
\includegraphics[width=0.98\textwidth]{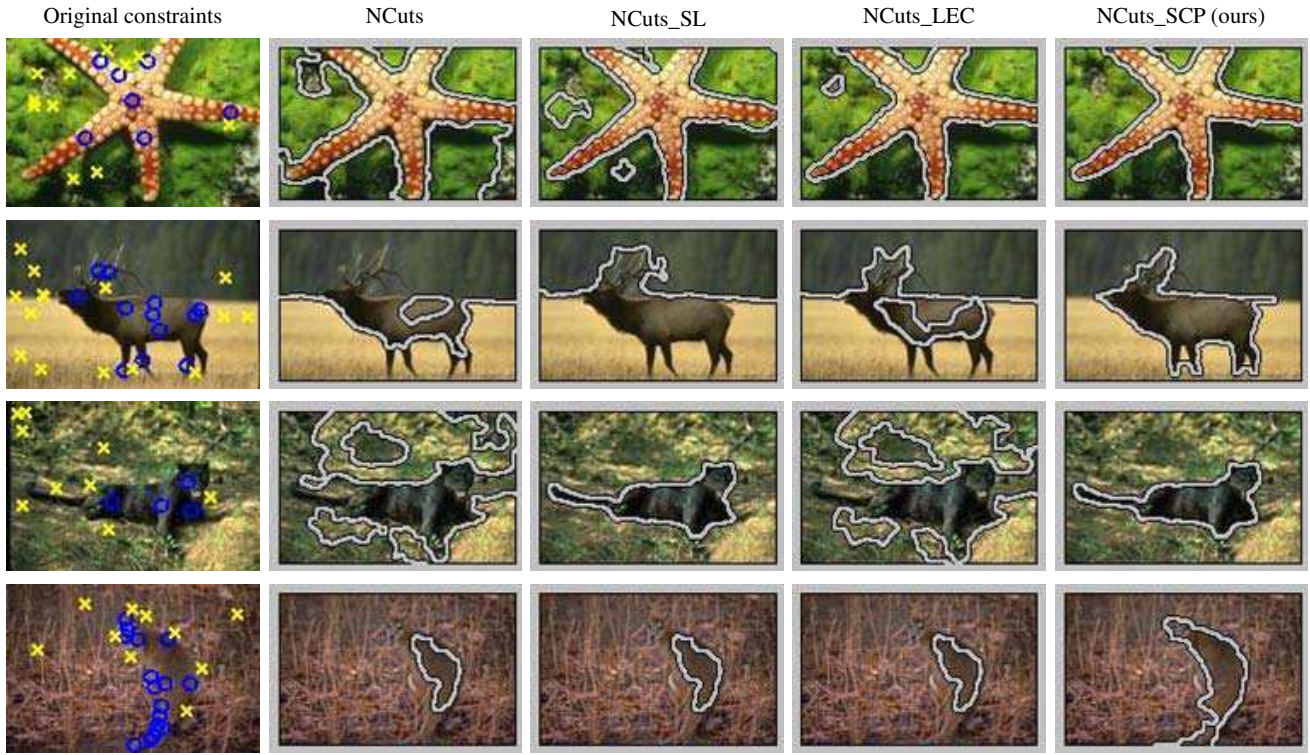}
\end{center}
\vspace{-0.1in} \caption{The results of constrained image segmentation
(cont.). A small set of labeled pixels (object and background denoted
by markers `o' and `x', respectively) are provided to infer the
initial set of linear equality constraints and pairwise constraints.}
\label{Fig.4} \vspace{-0.05in}
\end{figure*}

\subsection{Segmentation Results}

We first show the effect of $N_s$ (i.e. the number of randomly
selected pixels) on the performance of our NCuts\_SCP algorithm in
Table~\ref{Table.1}. Here, we measure the performance of our
NCuts\_SCP algorithm for constrained image segmentation by both AR
index and running time averaged over all the images. In particular, we
collect the running time by running our NCuts\_SCP algorithm (Matlab
code) on a computer with 3GHz CPU and 32GB RAM. From
Table~\ref{Table.1}, it can be clearly observed that our NCuts\_SCP
algorithm requires more running time but leads to higher AR index when
$N_s$ takes a larger value. Considering the tradeoff between the
effectiveness and efficiency of constrained image segmentation, we
select $N_s=2,500$ for our NCuts\_SCP algorithm. This setting is used
throughout the following experiments.

\begin{table}[t]
\vspace{-0.05in} \caption{The average segmentation results achieved by
different image segmentation methods.} \label{Table.2}
\vspace{-0.02in}
\begin{center}
\tabcolsep0.105cm
\begin{tabular}{|c|cccc|}
\hline
Methods     &  NCuts  & NCuts\_SL & NCuts\_LEC  &  NCuts\_SCP\\
\hline
AR index    &   0.36  & 0.39  &  0.40  &  0.50 \\
Time (sec.) &   28    &  31   &  25    &  35  \\
\hline
\end{tabular}
\end{center}
\vspace{-0.1in}
\end{table}

We further illustrate the selectively propagated constraints obtained
by our NCuts\_SCP algorithm with $N_s=2,500$ in Figure~\ref{Fig.3}.
Here, to explicitly represent the selectively propagated constraints,
we need to infer the labels of $N_s$ randomly selected pixels from
them. In fact, this inference can be done by simple voting according
to the output of our SCP with the initial set of labeled pixels being
regarded as the voters. Once we have inferred the labels of $N_s$
randomly selected pixels, we can show them out by marking the pixels
within object and background by blue `o' and yellow `x', respectively.
From Figure~\ref{Fig.3}, we find that the selectively propagated
constraints obtained by our NCuts\_SCP algorithm are mostly consistent
with the ground truth segmentation.

The comparison between different image segmentation methods is listed
Table~\ref{Table.2}. Meanwhile, this comparison is also illustrated in
Figures \ref{Fig.2} and \ref{Fig.4}. Here, the segmentation results
are evaluated by both AR index and running time averaged over all the
images. In particular, we collect the running time by running all the
algorithms (Matlab code) on a computer with 3GHz CPU and 32GB RAM. The
immediate observation is that our NCuts\_SCP algorithm significantly
outperforms the other three methods in terms of AR index. Since our
NCuts\_SCP algorithm incurs a time cost comparable to closely related
methods, it is preferred for constrained image segmentation by an
\emph{overall consideration}. In addition, it can be clearly observed that NCuts\_SCP, NCuts\_LEC, and NCuts\_SL lead to better results than the standard NCuts due to the use of constraints for image segmentation.

\section{Conclusions}
\label{sect:con}

In this paper, we have investigated the challenging problem of pairwise constraint propagation in constrained image segmentation. Considering the local homogeneousness of a natural image, we choose to
perform pairwise constraint propagation only over a selected subset of
pixels. Moreover, we solve such selective constraint propagation
problem by developing an efficient graph-based learning algorithm.
Finally, the selectively propagated constraints are used to adjust the
weight matrix based on $L_1$-minimization for image segmentation. The
experimental results have shown the promising performance of the
proposed algorithm for constrained image segmentation. For future
work, we will extend the proposed algorithm to other challenging tasks
such as semantic segmentation and multi-face tracking.

%

{\small
\bibliographystyle{named}

}

\end{document}